\documentclass[10pt,twocolumn,letterpaper]{article}

\usepackage[pagenumbers]{cvpr} 
\usepackage{bbding}
\usepackage[table]{xcolor}
\usepackage{tabularx} 
\usepackage{multirow} 
\definecolor{cvprblue}{rgb}{0.21,0.49,0.74}
\usepackage[pagebackref,breaklinks,colorlinks,allcolors=cvprblue]{hyperref}

\def\paperID{} 
\def\confName{CVPR}
\def\confYear{2026}

\title{RecurGS: Interactive Scene Modeling via Discrete-State\\ Recurrent Gaussian Fusion}

\author{
    Wenhao Hu\textsuperscript{1,2}\thanks{} \quad 
    Haonan Zhou\textsuperscript{1} \quad 
    Zesheng Li\textsuperscript{3} \quad 
    Liu Liu\textsuperscript{2} \quad 
    Jiacheng Dong\textsuperscript{1} \\
    Zhizhong Su\textsuperscript{2} \quad 
    Gaoang Wang\textsuperscript{1}\thanks{} \\ 
    \vspace{1em} 
    \textsuperscript{1}Zhejiang University \quad 
    \textsuperscript{2}Horizon Robotics \quad 
    \textsuperscript{3}Nanyang Technological University \\
    \vspace{0.5em}
}
\begin{document}
\twocolumn[{%
\renewcommand\twocolumn[1][]{#1}%
\maketitle
\vspace{-5em}
\begin{center}
    \centering
    \captionsetup{type=figure}
    \includegraphics[width=\textwidth, trim=0 0 0 0, clip]{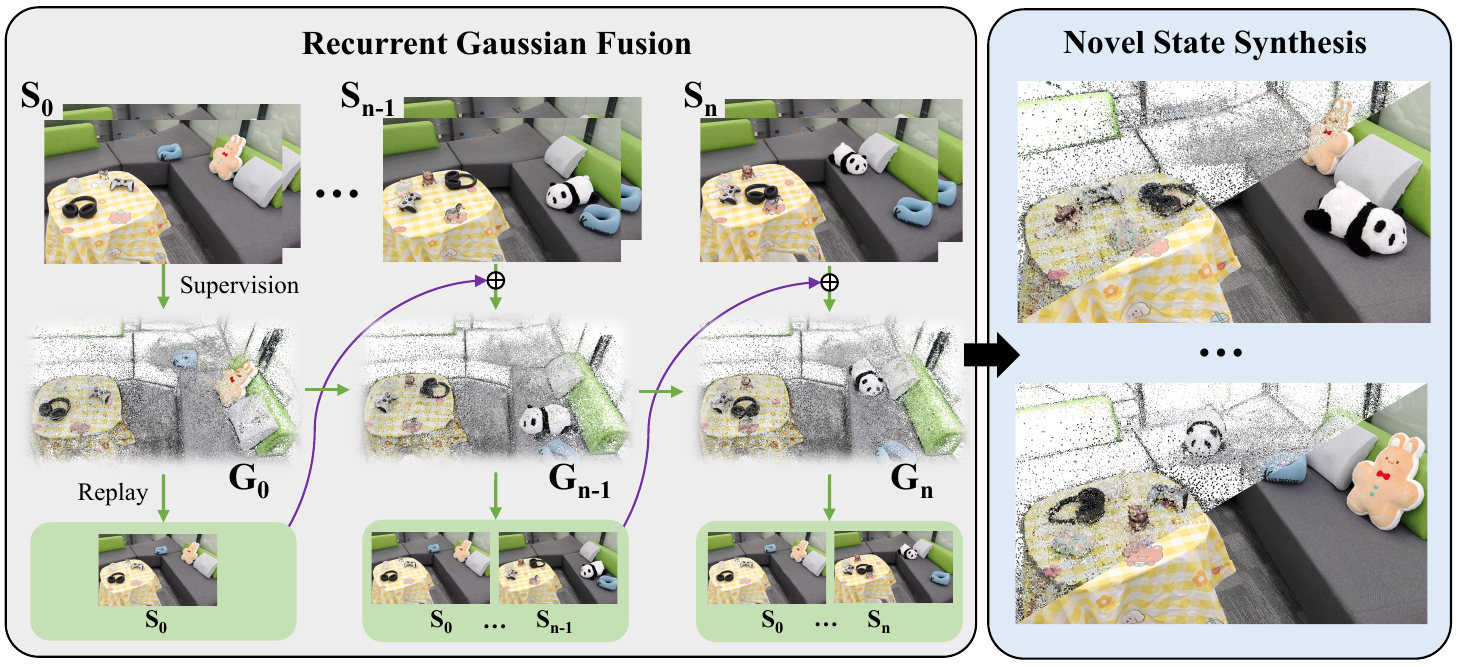}
    \captionof{figure}{Overview of our recurrent Gaussian fusion framework. 
Given a sequence of discrete scene states $S_0 \rightarrow S_n$, 
our model recurrently updates a single Gaussian scene representation 
by leveraging supervision from the current state and replay from past states. 
This produces a consistently fused scene $G_n$ that captures object-level 
changes across time, enabling controllable novel-state synthesis.
}
\label{fig:teaser}
\end{center}
}]

\footnotetext[1]{This work was done during an internship at Horizon Robotics.}
\footnotetext[2]{Corresponding author.}
\begin{abstract}
Recent advances in 3D scene representations have enabled high-fidelity novel view synthesis, yet adapting to discrete scene changes and constructing interactive 3D environments remain open challenges in vision and robotics. Existing approaches focus solely on updating a single scene without supporting novel-state synthesis. Others rely on diffusion-based object–background decoupling that works on one state at a time and cannot fuse information across multiple observations. To address these limitations, we introduce RecurGS, a recurrent fusion framework that incrementally integrates discrete Gaussian scene states into a single evolving representation capable of interaction. RecurGS detects object-level changes across consecutive states, aligns their geometric motion using semantic correspondence and Lie-algebra–based SE(3) refinement, and performs recurrent updates that preserve historical structures through replay supervision. A voxelized, visibility-aware fusion module selectively incorporates newly observed regions while keeping stable areas fixed, mitigating catastrophic forgetting and enabling efficient long-horizon updates. RecurGS supports object-level manipulation, synthesizes novel scene states without requiring additional scans, and maintains photorealistic fidelity across evolving environments. Extensive experiments across synthetic and real-world datasets demonstrate that our framework delivers high-quality reconstructions with substantially improved update efficiency, providing a scalable step toward continuously interactive Gaussian worlds.
\end{abstract}    
\section{Introduction}
\label{sec:intro}
Modeling interactive 3D scenes in discretely changing environments remains a 
fundamental yet underexplored problem in vision, robotics, and AR/VR~\cite{han2021reconstructing,rosinol20203d,wang2024architect,li2024scene,li2025radiance}. Real-world 
spaces evolve as objects are moved, added, or removed, creating a sequence of 
distinct scene states. A practical 3D representation must therefore integrate 
new observations over time, preserve previously learned structures, and retain 
object-level controllability for downstream interaction and reasoning~\cite{hu2025pointmap,booker2024embodiedrag,honerkamp2024language,wen20253d,li2024advances,li2025embodied}.

Existing efforts on temporal scene modeling primarily study continual learning 
for 3D reconstruction. Methods such as CL-NeRF~\cite{cai2023clnerf}, 
CL-Splats~\cite{ackermann2025cl}, and 
GaussianUpdate~\cite{zeng2025gaussianupdate} extend NeRF (neural radiance fields) or 3DGS (3D Gaussian Splatting) to sequential 
observations, improving update efficiency but focusing mainly on per-state 
reconstruction quality. These approaches neither capture structured transitions between states nor support object-aware manipulation.

\begin{figure}[!ht]
    \centering
    \includegraphics[width=1\linewidth]{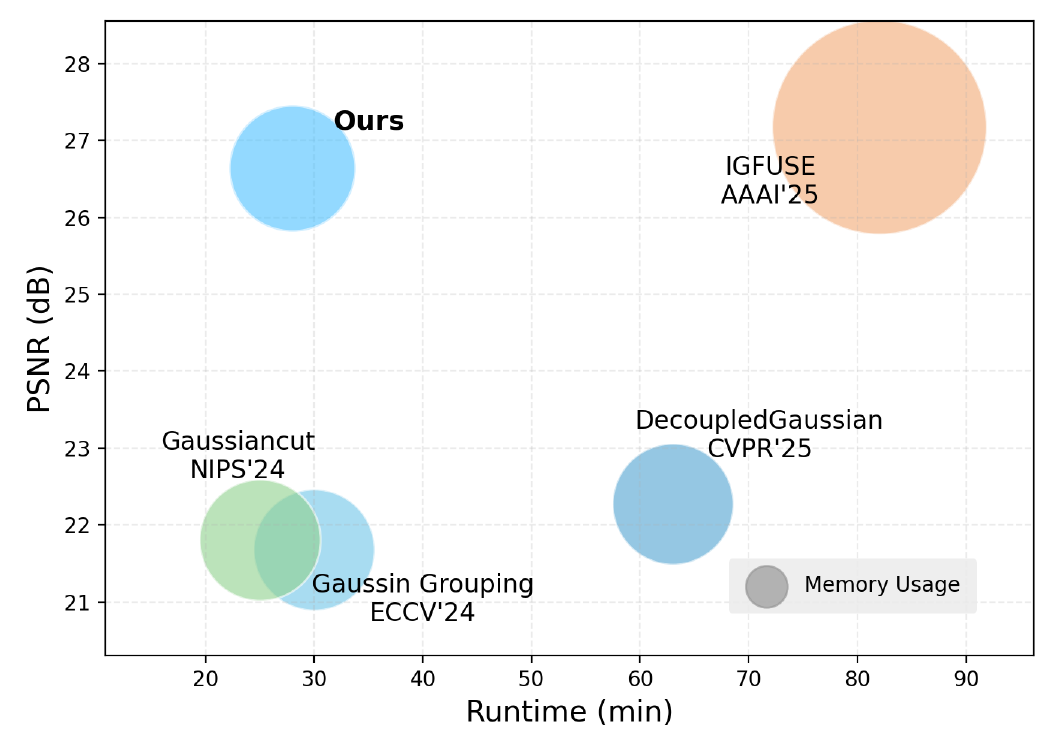}
    \caption{Efficiency vs. Quality. We compare RecurGS (Ours) against state-of-the-art methods on reconstruction quality, runtime, and GPU memory usage (bubble size). Unlike IGFuse which suffers from heavy memory overhead, our recurrent approach achieves comparable fidelity with significantly lower resource consumption.}
    \label{bubble}
\end{figure}

Another line of research investigates interactive scene representations through 
object decomposition or generative priors. Approaches including 
DecoupledGaussian~\cite{wang2025decoupledgaussian}, 
DP-Recon~\cite{ni2025decompositional}, and 
HoloScene~\cite{xia2025holoscene} enable editing or simulation by reconstructing 
object-centric scenes. However, they typically rely on diffusion-based priors or 
per-object optimization, resulting in heavy computation and limited adaptability 
to dynamically evolving real-world scenes. Consequently, current methods either 
update static representations without modeling multi-state evolution or build 
interactive scenes that are not designed for continual fusion across discrete 
states.
IGFuse~\cite{hu2025igfuseinteractive3dgaussian} fuses multi-scan observations for interactive Gaussian scene construction, but its all-at-once optimization requires loading and updating all scans simultaneously, resulting in heavy memory and time overhead and limiting scalability to long sequences or large numbers of scene states. Figure~\ref{bubble} quantitatively highlights this bottleneck, showing that prior methods consume excessive GPU memory or runtime compared to our efficient solution.

In this work, we introduce the task of \textbf{discrete-state recurrent 3D scene 
fusion} using 3D Gaussian Splatting. Our goal is to incrementally integrate 
observations from multiple scene states into a single evolving Gaussian model 
that can track object-level changes and support interaction. This enables the system to decompose, reconstruct, and manipulate real environments across multiple observations without requiring per-object dense scanning, providing a scalable solution for interactive and continually evolving 3D scene modeling. 

To address this task, we propose a recurrent fusion framework that maintains one 
Gaussian scene and updates it across discrete states through three stages. 
First, we analyze cross-state differences by detecting object-level changes and 
refining their geometric motion through semantic correspondence and Lie 
algebra–based SE(3) alignment. 
Second, we perform a recurrent update where the previous Gaussian scene implicitly encodes all past states and provides replay supervision to preserve historical structures.
Third, we selectively integrate new observations using a voxelized fusion module 
that updates only visible or newly revealed regions while keeping stable 
Gaussians fixed. 
Together, these steps produce a consistently evolving representation capable of 
long-horizon reasoning and object-level interaction. Our contributions are summarized as follows:
\begin{itemize}
\item We formulate discrete-state continual 3D scene fusion and introduce a 
framework that recurrently updates a single Gaussian scene without reinitialization.
\item We establish cross-state semantic–geometric consistency through CLIP-guided 
change detection and Lie algebra–based SE(3) pose refinement.
\item We propose a streaming voxel-based fusion strategy that updates only visible 
or newly revealed regions while preserving stable structures for efficiency and 
forgetting mitigation.
\item We enable object-level novel state synthesis by leveraging the emergent 
foreground–background separation within the recurrent Gaussian representation.
\end{itemize}

\section{Related Work}
\subsection{Continual Learning for Scene Reconstruction}
Continual learning (CL) for 3D scene reconstruction aims to incrementally update 
a scene representation as the environment evolves over time. Early point-cloud–based approaches such as RIO~\cite{wald2019rio}, 
Rescan~\cite{halber2019rescan}, and Living Scenes~\cite{zhu2024living} achieve object-level tracking and state matching, but their reliance on sparse point 
clouds results in limited rendering fidelity and weak supervision cues. To improve visual quality, recent research has shifted toward NeRF and Gaussian-based representations. NeRF-based continual learning methods—including CLNeRF~\cite{cai2023clnerf}, 
CL-NeRF~\cite{wu2023cl}, MEIL-NeRF~\cite{chung2025meil}, and 
IL-NeRF~\cite{zhang2025nerf}—commonly employ replay buffers, 
self-distillation, or lightweight adapters to incorporate new observations while 
alleviating catastrophic forgetting. In parallel, 3DGS-based approaches such as 
CL-Splats~\cite{ackermann2025cl} and GaussianUpdate~\cite{zeng2025gaussianupdate} 
propose efficient incremental frameworks that rapidly assimilate new views 
while preserving high-frequency geometric detail.

Despite these advances, existing CL frameworks remain limited to updating the
scene only at states that have been directly observed. They are unable to infer
or construct new scene configurations beyond the input observations, and they do
not reliably separate moving objects from the static background. These
constraints reduce their suitability for applications that require controllable,
interaction-aware, or simulation-ready scene understanding.

\subsection{Interactive Gaussian}
A growing body of work explores augmenting 3DGS reconstructions with simulation 
and editing capabilities. Several methods~\cite{liu2024physics3d,xie2024physgaussian,li2025pin} 
embed physical models into Gaussian scenes to enable forward simulation that 
adheres to real-world dynamics. RoboGSim~\cite{li2024robogsim} aligns reconstructed 
layouts with Isaac Sim to form a Real2Sim2Real loop for robotic evaluation, while 
Spring-Gaus~\cite{zhong2024reconstruction} integrates a mass–spring system with 
3DGS to jointly recover appearance, geometry, and deformation, supporting 
simulation under varying conditions. However, these approaches typically assume 
accurate scene or object scans, which limits their scalability in practice.

Another line of work focuses on data-driven generation and editing of interactive 
Gaussian scenes, leveraging strong priors from diffusion models. 
Decoupled Gaussians~\cite{wang2025decoupledgaussian} factorizes object and 
background components to enable independent manipulation, and 
GaussianEditor~\cite{chen2024gaussianeditor,wang2024gaussianeditor} provides 
fast, localized 3D editing through semantic tracing and hierarchical splatting. 
Recent systems such as HoloScene~\cite{xia2025holoscene} and 
DP-Recon~\cite{ni2025decompositional} reconstruct interactive, 
object-decomposed scenes with physical attributes or diffusion-based object 
priors. Although these approaches operate with limited observations and support 
rich interactions, they often rely on lengthy per-object optimization and can 
introduce geometric inconsistencies due to the use of generative priors.

The most related work is IGFuse~\cite{hu2025igfuseinteractive3dgaussian}, which also integrates multi-scan observations 
for interactive Gaussian scene construction. However, IGFuse jointly loads and 
optimizes all scans at once, resulting in high memory and time costs, whereas 
our recurrent fusion updates a single scene using only the current images, 
greatly improving scalability.

\section{Methodology}
\label{sec:method}

\subsection{Preliminaries}
We employ 3D Gaussian Splatting~\cite{kerbl3Dgaussians} to construct our 3D scenes, where the scene is represented as  a collection of 3D Gaussian primitives. Each Gaussian primitive is parameterized by a center point $\boldsymbol\mu$, a rotation matrix $\boldsymbol R$, a scaling matrix $\boldsymbol S$, an opacity $\boldsymbol \alpha$, and a color vector $\boldsymbol c$. To render a 2D image, 3DGS applies a differentiable $\alpha$-blending process that projects the 3D Gaussian primitives onto the image plane. The color of each pixel is computed by accumulating Gaussian contributions weighted by their projected opacities $\boldsymbol{\alpha}_i'$:
\begin{equation}
C = \sum_{i=1}^{N} \boldsymbol c_i {\boldsymbol \alpha}_i '\prod_{j=1}^{i-1}(1-{\boldsymbol \alpha}_j')
\end{equation}

\subsection{Task Formulation}
\begin{figure*}[t]
  \centering
   \includegraphics[width=\linewidth]{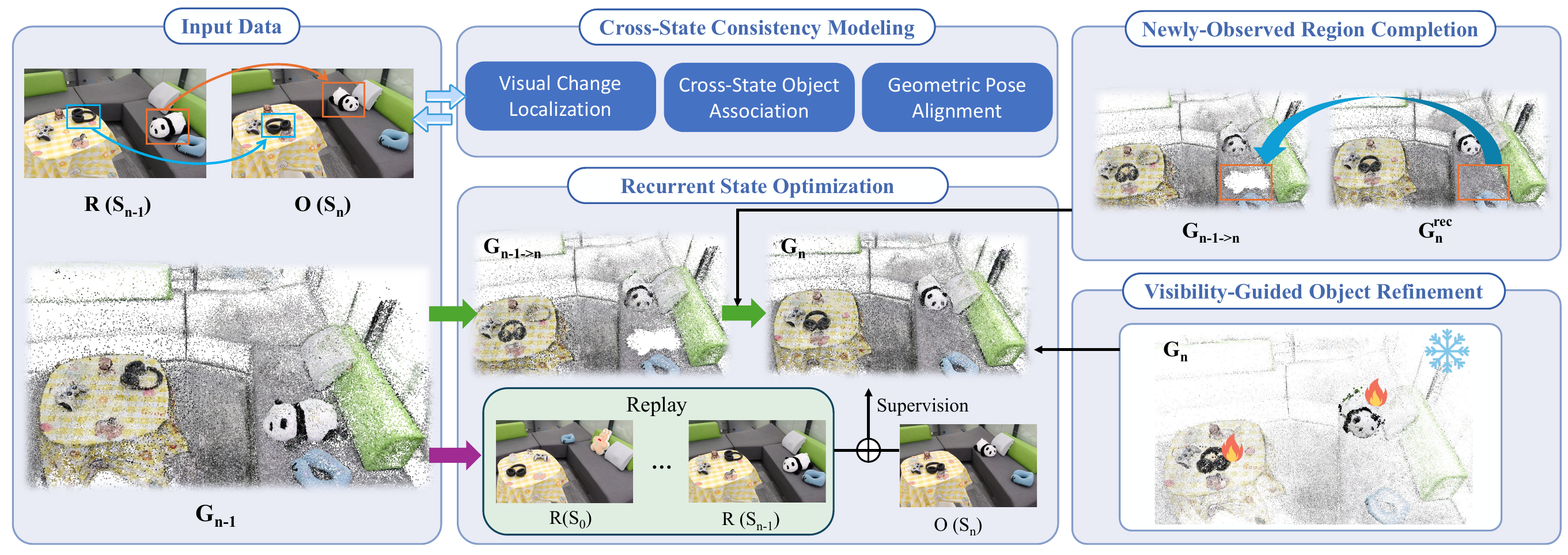}
   \caption{Overview of our recurrent Gaussian scene fusion pipeline.
Given two consecutive scene states, the system first performs cross-state consistency modeling through visual change localization, object association, and geometric pose alignment. The previous Gaussian scene is then recurrently optimized using supervision from the current state and replayed renderings of earlier states. Newly observed regions are completed using 3D initialization from a separately reconstructed scene, and visibility-guided refinement updates only object-relevant Gaussians, resulting in an efficient and consistent fused representation. Here, $R$ denotes rendered images from the Gaussian scene and
$O$ denotes the observed input images.
}
   \label{fig:framework}
\end{figure*}
We address the task of interactive scene modeling through continual learning of temporally evolving environments, where the scene evolution is represented as a sequence of discrete time steps. 
\paragraph{Scene update}
At the initial timestamp $t{=}0$, we reconstruct a base Gaussian scene 
$\mathcal{G}_{\text{rec}}^{0}$ from a set of multi-view observations 
$\mathcal{I}^0=\{I_i^0\,|\,i=1,\ldots,n\}$ using standard 3D Gaussian Splatting optimization. 
This reconstructed scene serves as the initial state of the system and is continuously maintained throughout the sequence.

At each subsequent time step $t$, given new multi-view observations 
$\mathcal{I}^t=\{I_i^t\,|\,i=1,\ldots,n\}$ from n views and the previous Gaussian state $\mathcal{G}^{t-1}$, which compactly encodes all accumulated information from time steps $0$ to ${t-1}$, our goal is to update the scene to its new state $\mathcal{G}^t$. 
To achieve this, we introduce a state-transition operator 
$\mathcal{S}_{t-1\rightarrow t}$ that governs the evolution of the scene representation from $\mathcal{G}^{t-1}$ to $\mathcal{G}^t$. 
Unlike purely geometric transformations, this operator encapsulates the entire update process, including change detection, alignment, and recurrent optimization, enabling continual scene refinement without re-initialization.

Formally, the recursive update of the Gaussian scene is defined as:
\begin{equation}
\mathcal{G}^t =
\mathcal{S}_{t-1\rightarrow t}\!\left(
    \mathcal{G}^{t-1},\, \mathcal{I}^t
\right),
\quad t = 1,2,\ldots,T
\end{equation}
where $\mathcal{G}^0 = \mathcal{G}_{\text{rec}}^{0}$. 
This recursive formulation allows the system to maintain a single unified Gaussian scene that evolves consistently over time, 
continuously integrating new observations and supporting object-level interaction within a dynamically changing environment.

\paragraph{Novel State Synthesis}
During the continual update process, the evolving Gaussian scene $\mathcal{G}^t$ naturally separates stable background regions from manipulable foreground structures. 
Leveraging this property, we enable novel state synthesis by directly applying an object-level transformation matrix $\mathbf{T} \in SE(3)$ to the foreground Gaussians, yielding a new hypothetical scene configuration:
\begin{equation}
\mathcal{G}^{\mathrm{novel}} =
\mathbf{T}\!\left(\mathcal{G}^{t}\right)
\end{equation}
where only the foreground components affected by $\mathbf{T}$ are spatially relocated, while the background remains unchanged. 
This mechanism allows the system to generate new scene states without additional observations, supporting downstream tasks such as interactive editing, counterfactual reasoning, and robot-driven environment manipulation.

\subsection{Cross-State Consistency Modeling}

To compute the scene-state transition $\mathcal{S}_{t-1\rightarrow t}$,
we first establish cross-state consistency in semantics and geometry pose between consecutive scene states.

\paragraph{Visual Change Localization.}
Given new observations $\mathcal{I}^t$, we render the previous state $\mathcal{G}^{t-1}$ under the current camera poses,
obtaining $\tilde{\mathcal{I}}^{t-1}$.
Paired images $\{\tilde{\mathcal{I}}^{t-1}, \mathcal{I}^t\}$ allow direct comparison under identical viewpoints.
We extract DINOv3~\cite{simeoni2025dinov3} features and compute pixel-wise cosine similarity:
\begin{equation}
\mathcal{F}_{\text{DINO}}^{t-1} = \mathcal{E}_{\text{DINO}}(\tilde{\mathcal{I}}^{t-1}),\quad
\mathcal{F}_{\text{DINO}}^{t} = \mathcal{E}_{\text{DINO}}({\mathcal{I}}^t)
\end{equation}
\begin{equation}
\mathcal{M}_{\text{DINO}}^{t} =
\cos\!\left(\mathcal{F}_{\text{DINO}}^{t-1},\mathcal{F}_{\text{DINO}}^t\right)
\end{equation}
yielding a coarse region-of-change mask.
We further refine object-level masks by intersecting SAM2~\cite{ravi2024sam} segmentations with the change mask across multiple views, yielding the multi-view object masks $\mathcal{M}_{\text{Obj}}^{t-1}$ and $\mathcal{M}_{\text{Obj}}^{t}$ .

\paragraph{Cross-State Object Association.}
To determine whether each changed region corresponds to a removed, added, or moved object,
we extract CLIP~\cite{radford2021learning} features for the refined masks:
\begin{equation}
\mathcal{F}_{\text{CLIP}}^{t-1} =
\mathcal{E}_{\text{CLIP}}(\tilde{\mathcal{I}}^{t-1}\!\cdot\!\mathcal{M}_{\text{Obj}}^{t-1}), \quad
\mathcal{F}_{\text{CLIP}}^{t} =
\mathcal{E}_{\text{CLIP}}({\mathcal{I}}^{t}\!\cdot\!\mathcal{M}_{\text{Obj}}^t)
\end{equation}
One-to-one matching via the Hungarian algorithm aligns semantic identities across states.
Unmatched elements in $\mathcal{F}_{\text{CLIP}}^{t-1}$ or $\mathcal{F}_{\text{CLIP}}^{t}$ indicate removed or newly added objects.

\paragraph{Geometric Pose Alignment.}
Using the refined masks, we segment the corresponding Gaussian primitives~\cite{jain2024gaussiancut}
and obtain object subsets $\{\mathcal{O}^{t-1},\mathcal{O}^{t}\}$.
For moving objects, we first estimate a coarse pose via ICP~\cite{zhang2021fast}, producing 
$T_{\text{coarse}} \in SE(3)$
To obtain a more accurate alignment, we optimize the Lie-algebra parameter 
$\xi \in \mathfrak{se}(3)$ of the rigid transform while keeping all Gaussian 
parameters fixed. The refined pose is obtained by solving
\begin{equation}
\xi^{*} = \arg\min_{\xi}\,\mathcal{L}_{\text{align}}(\xi)
\end{equation}
where $\mathcal{L}_{\text{align}}(\xi)$ measures photometric and geometric 
consistency between the transformed $\mathcal{G}^{t-1}$ and the observations at time $t$.
The SE(3) transformation corresponding to $\xi$ is recovered through the 
exponential map $\operatorname{Exp}(\xi)=\exp([\xi]_\times)$, and the 
refined transformation is therefore
\begin{equation}
T_{\text{fine}} = \operatorname{Exp}(\xi^{*})
\end{equation}

Finally, the updated Gaussian state is obtained by applying this transform:
\begin{equation}
\mathcal{G}^{t-1\to t} = T_{\text{fine}}(\mathcal{G}^{t-1})
\end{equation}
This transformed state may contain empty or incomplete background regions and thus requires further completion and refinement.
\subsection{Recurrent State Fusion}

After estimating the object motion and obtaining the updated scene 
$\mathcal{G}^{t-1\rightarrow t}$, the system must integrate the new observations 
$\mathcal{I}^t$ into the recurrent Gaussian representation. 
A key property of our formulation is that $\mathcal{G}^{t-1}$ implicitly carries 
all accumulated information from previous steps $\{0,\ldots,t-1\}$, enabling 
replay-based supervision and preventing catastrophic forgetting. 
To efficiently update only the regions supported by new evidence, 
we adopt a voxelized spatial index that enables fast visibility reasoning and 
new-region selection. 
\paragraph{Visibility-Guided Object Refinement}
\begin{figure}[!t]
    \centering
    \includegraphics[width=1\linewidth]{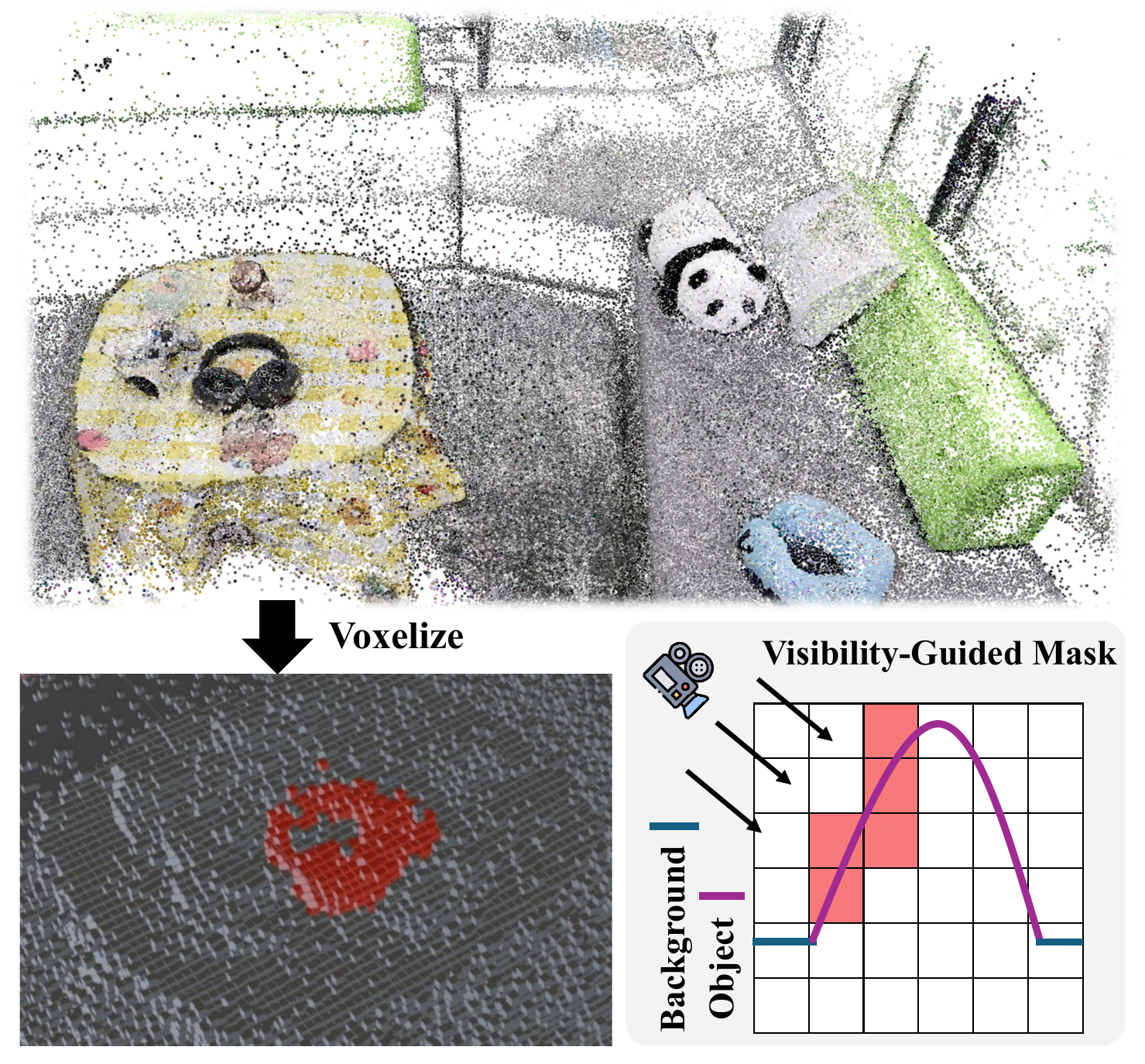}
    \caption{Our method updates only the regions corresponding to moving objects, improving optimization efficiency, while unseen areas are masked out to prevent unintended updates and forgetting of previously reconstructed content.}
    \label{vis_mask}
\end{figure}
As shown in Figure~\ref{vis_mask}, we refine the portions of moving objects that are visible to the cameras 
at time $t$. Using the estimated transformation $T_{\text{fine}}$, we map 
the object $\mathcal{O}^{t-1}$ into its current pose and perform ray tracing 
over the object voxel grid $V_{\text{obj}}^{t-1 \to t}$ under the new camera views. 
The ray-tracing process produces a per-voxel visibility mask 
$\text{Mask}_{\text{vis}}$ that marks voxels hit by at least one primary ray, 
thereby indicating the object regions falling inside the camera frustums. 
We then gather the Gaussian primitives whose hosting voxels satisfy 
$\text{Mask}_{\text{vis}}=1$, forming the visible subset 
$\mathcal{G}_{\text{vis}}$ used for refinement. 
Restricting optimization to this visible region prevents overfitting to 
unobserved surfaces and ensures that updates occur only where photometric 
evidence is available.

\paragraph{Newly-Observed Region Completion}
Scene changes often expose regions that were never observed in earlier states, 
including both disoccluded background and newly appearing foreground surfaces. 
To complete these newly observed areas, we voxelize the reconstructed scene 
$\mathcal{G}^t_{\mathrm{rec}}$ into $V^t$ and the previous object region 
into $V_{\text{obj}}^{t-1}$. 
We identify two types of regions requiring initialization: 
(1) background that becomes visible after an object moves away, captured by the 
intersection $V^t \cap V_{\text{obj}}^{t-1}$, and 
(2) newly appearing foreground surfaces or entirely new objects, represented by 
the difference $V^t \setminus V_{\text{obj}}^{t-1}$. 
The union of these regions forms the fill set,
\begin{equation}
    V_{\text{fill}} = (V^t \cap V_{\text{obj}}^{t-1}) \;\cup\; (V^t \setminus V_{\text{obj}}^{t-1})
\end{equation}
from which we extract the corresponding Gaussian primitives 
$\mathcal{G}_{\text{fill}}$ and insert them into the recurrent representation. 
This initialization provides physically consistent geometry and appearance for 
regions that have never been observed before, and accelerates their convergence during optimization.

\paragraph{Recurrent State Optimization}
We jointly optimize the Gaussian subsets 
$\mathcal{G}_{\text{vis}}$ (visible object regions) and 
$\mathcal{G}_{\text{fill}}$ (newly observed regions) within the recurrent scene 
$\mathcal{G}^{t-1\rightarrow t}$. 
To maintain temporal coherence, the updated scene is supervised by two 
complementary signals. 

First, we leverage the historical information preserved in 
$\mathcal{G}^{t-1}$ through a random replay mechanism. 
At each optimization step, we uniformly sample a past time index 
$i \sim \mathcal{U}(0,t-1)$ and render a pseudo ground-truth image 
$I_{\text{replay}}^{i}$ from $\mathcal{G}^{t-1}$. 
This replay image supervises the updated scene 
$\mathcal{G}^{t-1\rightarrow t}$, whose rendering at time $i$ is denoted as 
$I_{\text{render}}^{i}$. 
The replay loss is defined as:
\begin{equation}
\mathcal{L}_{\text{replay}}
= (1-\lambda_{s})\,
\mathcal{L}_{\text{L1}}(I_{\text{render}}^{i}, I_{\text{replay}}^{i})
+ \lambda_{s}\,
\mathcal{L}_{\text{SSIM}}(I_{\text{render}}^{i}, I_{\text{replay}}^{i})
\end{equation}

Second, we supervise the same updated scene using the ground-truth observations 
at the current time step~$t$. 
Let $I_{\text{render}}^{t}$ denote the rendering of $\mathcal{G}^{t-1\rightarrow t}$ 
under the current camera views. The current-state loss is:
\begin{equation}
\mathcal{L}_{\text{curr}}
= (1-\lambda_{s})\,
\mathcal{L}_{\text{L1}}(I_{\text{render}}^{t}, I_{\text{gt}}^{t})
+ \lambda_{s}\,
\mathcal{L}_{\text{SSIM}}(I_{\text{render}}^{t}, I_{\text{gt}}^{t})
\end{equation}

The final optimization objective combines both components:
\begin{equation}
\mathcal{L} = \lambda_{r}\mathcal{L}_{\text{replay}} + (1-\lambda_{r})\mathcal{L}_{\text{curr}}
\end{equation}
jointly enforcing consistency with past states and fidelity to new observations, 
leading to stable and temporally coherent incremental scene refinement.

\begin{figure*}[!t]
    \centering
    \includegraphics[width=1\linewidth]{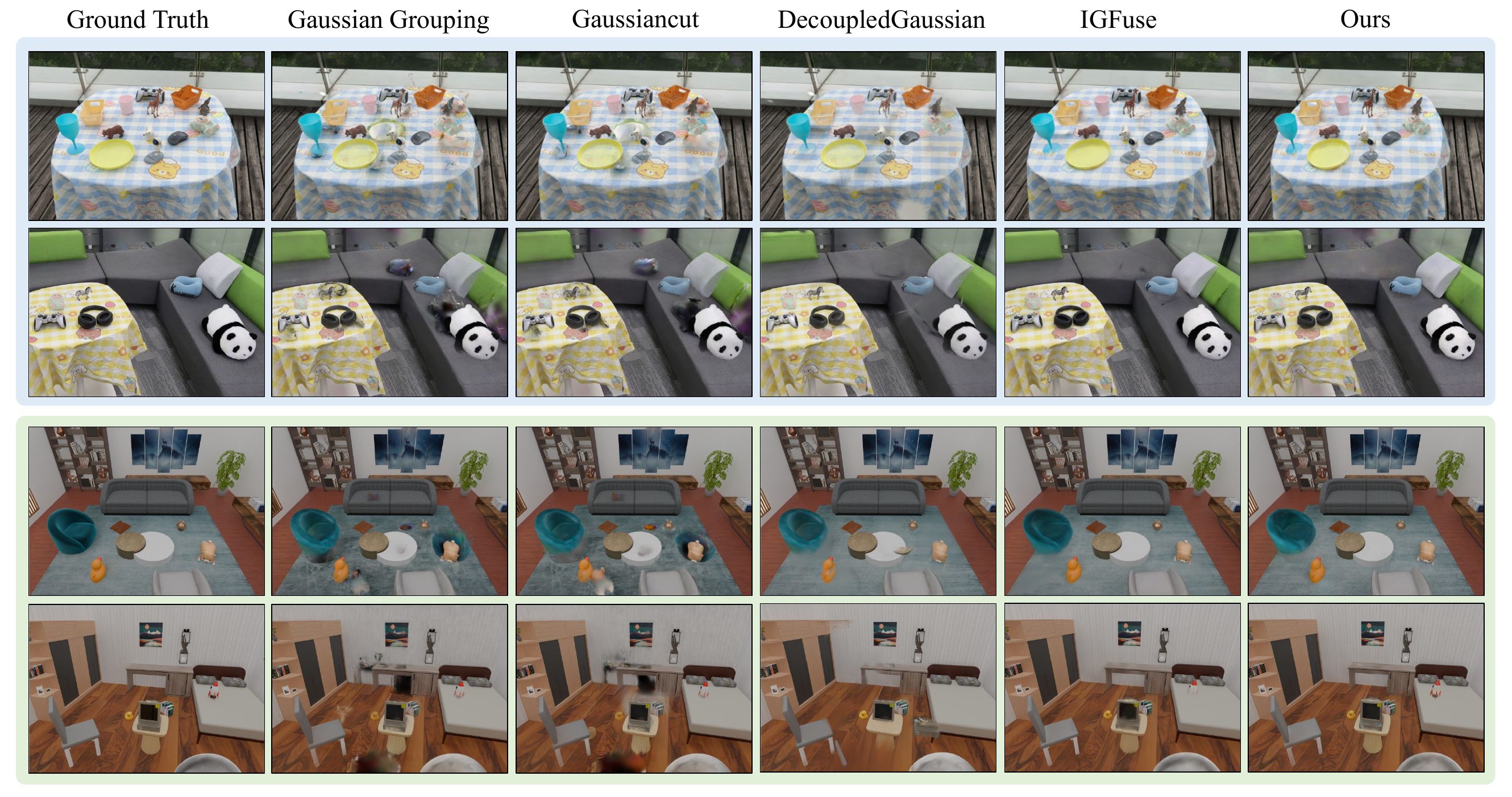}
    \caption{Qualitative comparison of novel-state synthesis on real-world (top) and synthetic (bottom) scenes. While existing methods exhibit boundary artifacts, missing background regions, or object mixing, our approach produces accurate and complete novel-state renderings that closely match the ground truth.}

    \label{novel}
\end{figure*}
\section{Experiments}
\label{sec:exp}

\subsection{Dataset}
\label{dataset}

Our primary experiments are conducted on the multi-scan dataset proposed in
IGFuse~\cite{hu2025igfuseinteractive3dgaussian}, which contains 7 synthetic scenes created in
Blender~\cite{blender} with textured assets from BlenderKit~\cite{blenderkit},
and 5 real-world scenes captured using handheld RGB cameras. Each scene provides
multiple interaction states obtained by changing object poses between scans, and
a separate test configuration for evaluating object-level rearrangement. To further study long-horizon fusion efficiency, we extend the IGFuse dataset by increasing the number of training states from the original 2--3 to 6--8 per scene. This produces longer and more
challenging state sequences. The evaluation protocol follows IGFuse, where a test state
is created by random object repositioning and PSNR/SSIM are computed on
predefined camera views. Additional dataset construction details are provided in the appendix.

\subsection{Experimental Setup}
\paragraph{Implementation details}
During pose refinement, we perform 1,000 iterations of Lie-algebra optimization to estimate the fine-grained SE(3) transformation. For voxelized fusion, we discretize the Gaussian field with a voxel size of 0.05\,m, which provides an efficient spatial index for visibility reasoning and region selection. During recurrent optimization, we fine-tune the updated Gaussian scene for 2,000 iterations. For the reconstruction losses, we set the SSIM weight to $\lambda_s = 0.2$ and the replay loss weight to $\lambda_r = 0.5$. All experiments are trained on a single NVIDIA RTX~4090 GPU.
\paragraph{Baselines}
We compare our method with representative Gaussian Splatting–based scene modeling approaches. Gaussian Grouping~\cite{ye2024gaussian} performs feature-based clustering of Gaussians. GaussianCut~\cite{jain2024gaussiancut} solves a graph-cut formulation over Gaussian primitives. Decoupled Gaussian~\cite{wang2025decoupledgaussian} further incorporates remeshing and LaMa-based inpainting to obtain more complete object geometry. IGFuse~\cite{hu2025igfuseinteractive3dgaussian} performs multi-scan fusion by loading all scans at once and jointly optimizing all Gaussian fields in a single global optimization stage.

\subsection{Novel State Synthesis}
\begin{table}[!t]
\caption{Quantitative comparison of novel state synthesis quality on the \textbf{synthetic} dataset.}
\label{tab:2d_data_synthetic}
\centering
\begin{tabular}{l|c|c|c}
\toprule
\textbf{Model} & PSNR & SSIM & Time(min) \\
\midrule
Gaussian Grouping~\cite{ye2024gaussian}  
    & 28.93 
    & 0.950 
    & 27 \\

Gaussiancut~\cite{jain2024gaussiancut}  
    & 29.01 
    & 0.956 
    & \cellcolor{blue!40}20 \\

DecoupledGaussian~\cite{wang2025decoupledgaussian} 
    & 30.27 
    & 0.959 
    & 55 \\

IGFuse~\cite{hu2025igfuseinteractive3dgaussian} 
    & \cellcolor{blue!40}36.93
    & \cellcolor{blue!40}{0.978}
    & 75 \\

\textbf{Ours} 
    & \cellcolor{blue!15}36.61
    & \cellcolor{blue!15}0.977
    & \cellcolor{blue!15}23 \\
\bottomrule
\end{tabular}
\end{table}

\begin{table}[!t]
\caption{Quantitative comparison of novel state synthesis quality on the \textbf{real-world} dataset.}
\label{tab:2d_data_real}
\centering
\begin{tabular}{l|c|c|c}
\toprule
\textbf{Model} & PSNR & SSIM & Time(min) \\
\midrule

Gaussian Grouping~\cite{ye2024gaussian}  
    & 21.68 & 0.853 & 30 \\

Gaussiancut~\cite{jain2024gaussiancut}  
    & 21.81 & 0.864 & \cellcolor{blue!40}25 \\

DecoupledGaussian~\cite{wang2025decoupledgaussian} 
    & 22.28 & 0.855 & 63 \\

IGFuse~\cite{hu2025igfuseinteractive3dgaussian} 
    & \cellcolor{blue!40}27.18 
    & \cellcolor{blue!40}0.907
    & 82 \\

\textbf{Ours} 
    & \cellcolor{blue!15}26.64
    & \cellcolor{blue!15}0.898
    & \cellcolor{blue!15}28 \\
\bottomrule
\end{tabular}

\end{table}

We present qualitative comparisons on the real-world dataset (Figure.~\ref{novel}, top) and the synthetic dataset (Figure.~\ref{novel}, bottom). Methods relying solely on segmentation struggle to maintain completeness during novel-state synthesis. Even when trained with multiple scans, segmentation-based methods suffer from the inherent overfitting tendency of Gaussian Splatting, causing previously learned information to be forgotten and leading to noticeable holes in the background when objects move to new states. Among them, the graph-based Graussiancut suppresses boundary residuals more effectively than the feature-based Gaussian Grouping, yet substantial artifacts still remain. DecoupledGaussian, which incorporates diffusion priors alongside background Gaussian completion, 2D inpainting, and Gaussian fine-tuning, generates visually coherent results on synthetic scenes where backgrounds are simple and structured. However, its LaMa-based inpainting module fails to reliably restore fine-grained details in real scenes with more complex and cluttered backgrounds. Both IGFuse and our method effectively leverage complementary information across multiple scene states, achieving image synthesis results that are nearly indistinguishable from ground truth. 

\begin{figure*}[!t]
    \centering
    \includegraphics[width=1\linewidth]{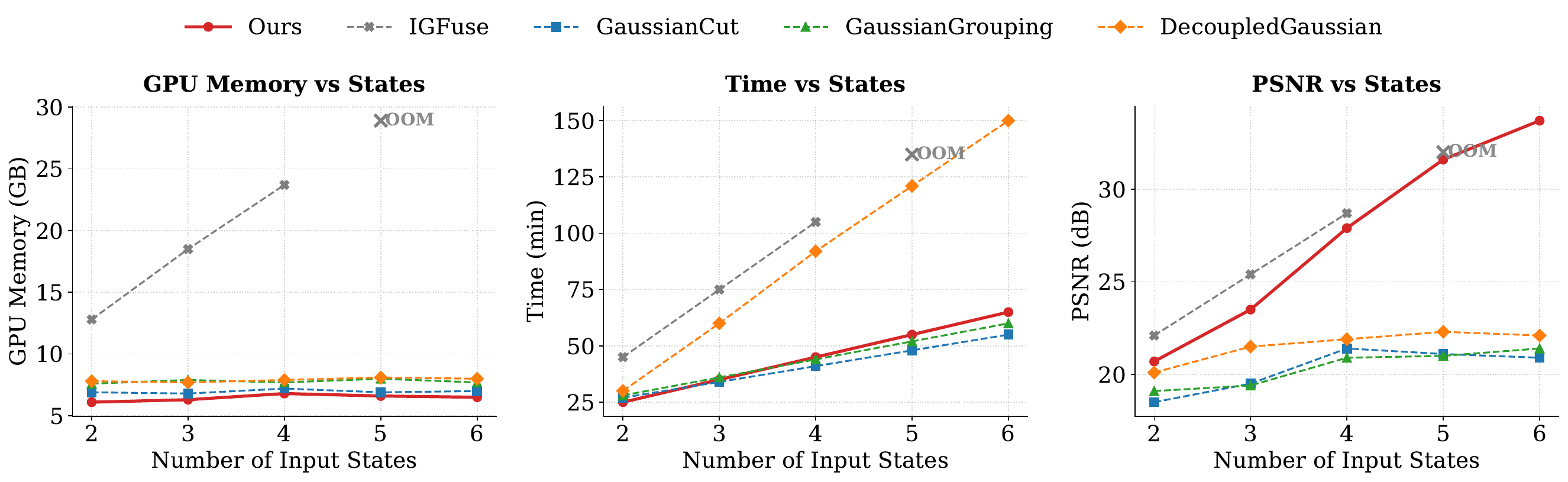}
    \caption{Memory, runtime, and PSNR vs. number of input states. Our method scales efficiently in both memory and time, and continues improving in quality, whereas competing methods grow costly and show saturated PSNR.}
    \label{time}
\end{figure*}

As seen in Table~\ref{tab:2d_data_synthetic} and Table~\ref{tab:2d_data_real}, methods that rely solely on segmentation achieve relatively low PSNR and SSIM. Incorporating inpainting modules improves performance, with particularly strong gains on synthetic scenes where the backgrounds are simpler and more regular. In contrast, the improvements on real-world scenes are less pronounced due to the higher complexity and richer textures of real backgrounds. By leveraging complementary information across multiple scene states, both IGFuse and our method outperform the other baselines. Our PSNR is slightly lower than IGFuse because it uses pseudo-state supervision, which is particularly beneficial when only a few training states are available. More importantly, unlike IGFuse which jointly optimizes all scenes and images in a single batch, our recurrent fusion design achieves 3× faster optimization while maintaining highly competitive reconstruction quality.

\subsection{Time Efficiency Comparison}
\label{tec}
\begin{figure*}[!t]
    \centering
    \includegraphics[width=1\linewidth]{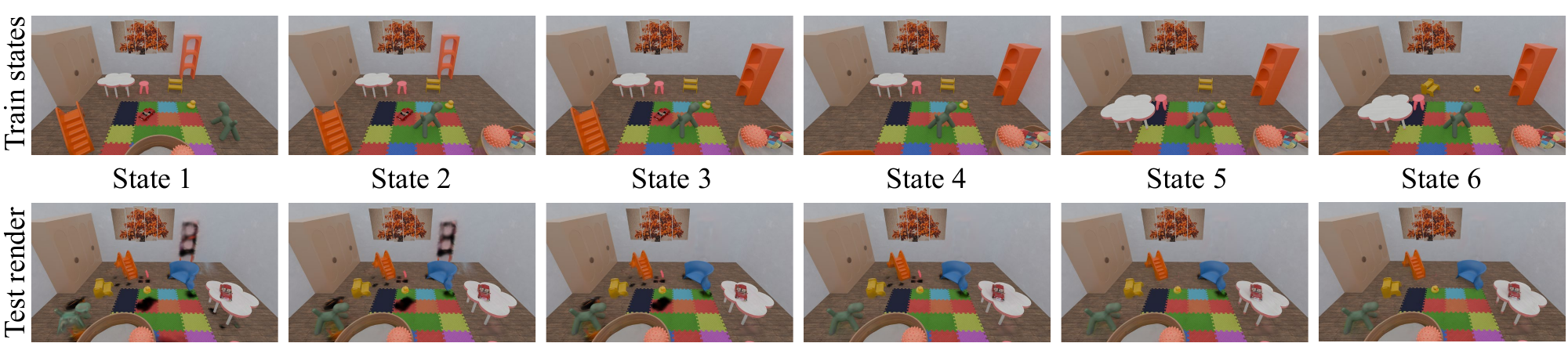}
    \caption{Visualization of incremental scene refinement. The top row shows training states ($S_1 \rightarrow S_6$) where object rearrangement exposes previously occluded regions. The bottom row displays the corresponding test renderings. As the system fuses more states, it integrates information from newly revealed areas, progressively eliminating artifacts and improving scene completeness.}
    \label{long_main}
\end{figure*}
Figure~\ref{time} shows how computational cost scales with the number of scene states. Because IGFuse jointly loads and optimizes all Gaussian scenes at once, its GPU memory usage grows rapidly and runs out of memory when four input states are provided. In contrast, our recurrent fusion strategy maintains a nearly constant memory footprint as the number of states increases. A similar trend appears in runtime: IGFuse requires significantly more time than other methods, while diffusion-prior–based methods such as DecoupledGaussian also incur high cost due to LaMa inpainting and subsequent finetuning.

In terms of reconstruction quality, IGFuse initially achieves higher PSNR because of its pseudo-state supervision. However, as more states are introduced, the number of pseudo states naturally becomes limited. Meanwhile, other baselines fail to improve over time since they do not retain information from previous states, resulting in stagnated PSNR across additional inputs.

In contrast, RecurGS effectively leverages the increasing number of states to accumulate complementary visual evidence. This benefit is qualitatively demonstrated in Figure~\ref{long_main}, which shows that our method progressively fills in missing background geometry and eliminates artifacts as the sequence evolves. As objects move across the training states (Top), previously occluded background regions are progressively dis-occluded. Early reconstructions suffer from missing visual evidence, resulting in visible artifacts (Bottom, left). However, as RecurGS incrementally fuses more states, it integrates these complementary observations to accurately fill in missing geometry. This continuous refinement effectively eliminates artifacts and ensures high-fidelity reconstruction without relying on generative hallucinations.

\subsection{Ablation Studies}
\label{sec:ablation}
Table~\ref{tab:ablation} summarizes the ablation study on the effectiveness of each component in our framework: R (Recurrent State Fusion), N (Newly-Observed Region Completion), and V (Visibility-Guided Object Refinement). Using only Recurrent State Fusion already provides a strong baseline by continuously integrating information from multiple scene states, achieving a PSNR of 35.84. Adding Newly-Observed Region Completion further improves reconstruction quality, as the model can directly obtain 3D initialization for new or previously unseen regions from the pre-reconstructed scene. This approach is both more efficient and more effective than optimizing hole regions through split and clone operations. Finally, Visibility-Guided Object Refinement restricts updates to only the moving objects while keeping the static background fixed, which significantly reduces optimization time with negligible impact on PSNR.
\begin{table}[t]
  \caption{Ablation study of R (Recurrent State Fusion), N (Newly-Observed Region Completion), and V (Visibility-Guided Object Refinement).}
  \label{tab:ablation}
  \centering
  \renewcommand{\arraystretch}{0.9}
  \begin{tabular*}{\columnwidth}{@{\extracolsep{\fill}}ccc|ccc}
    \toprule
    R & N & V & PSNR~$\uparrow$ & SSIM~$\uparrow$ & Time (min)~$\downarrow$ \\
    \midrule
    \CheckmarkBold & - & - & 35.84 & 0.965 & 30 \\
    \CheckmarkBold & \CheckmarkBold & - & 36.58 & 0.975 & 30 \\
    \CheckmarkBold & \CheckmarkBold & \CheckmarkBold & \textbf{36.61} & \textbf{0.977} & \textbf{23} \\
    \bottomrule   
  \end{tabular*}
\end{table}

\begin{figure}[!t]
    \centering
    \includegraphics[width=1\linewidth]{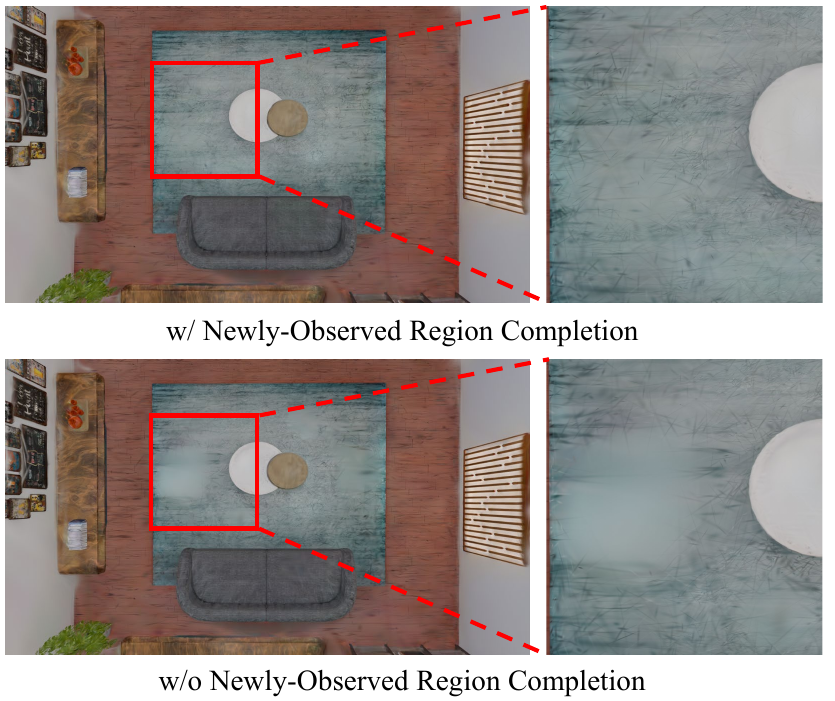}
    \caption{Effect of Newly-Observed Region Completion. With region completion, newly revealed background areas are accurately reconstructed, whereas removing this component leads to blurry result.}
    \label{completion}
\end{figure}

Figure~\ref{completion} further highlights the impact of Newly-Observed Region Completion. When this component is enabled, the model can directly initialize newly revealed regions from the pre-reconstructed scene, producing clean and consistent background geometry. In contrast, without this module, the system must rely solely on split-and-clone optimization, which is insufficient to repair large missing areas within a short finetuning window, leading to blurred textures and incomplete background filling. This demonstrates that proper 3D initialization is essential for maintaining background coherence during state transitions.

\subsection{Pose Alignment}
Table~\ref{tab:pose_psnr} demonstrates that our two-stage pose estimation pipeline using ICP followed by RT-Optimization is highly effective. In this setting, we evaluate on the synthetic dataset where Blender provides accurate ground-truth camera poses, and PSNR is computed only over the object regions. The optimized poses yield a PSNR that is nearly identical to that obtained with ground-truth poses, and RT-Optimization consistently improves the coarse alignment provided by ICP.

Table~\ref{tab:pose_noise} further evaluates robustness by injecting noise into the ICP-estimated transformations across both rotation and translation dimensions. The refined results show a clear recovery of PSNR under all noise levels, indicating that even when ICP provides an imperfect initial alignment, our optimization stage is able to correct the pose and restore high-quality reconstructions.

\begin{table}[t]
    \caption{Comparison of PSNR under different pose estimation strategies.}
    \label{tab:pose_psnr}
    \centering
    \begin{tabularx}{\linewidth}{lXXX}
        \toprule
        \textbf{Metric} & \textbf{$T_{\text{coarse}}$} & \textbf{$T_{\text{fine}}$}& \textbf{GT Pose} \\
        \midrule
        PSNR & 39.29 & 39.37 & 39.45 \\
        \bottomrule
    \end{tabularx}
   
\end{table}

\begin{table}[t]
    \caption{Initial vs refined performance under different pose noise levels.}
    \label{tab:pose_noise}
    \centering
    \begin{tabular}{l|cc|cc}
        \toprule
        \multirow{2}{*}{\textbf{Noise}} &
        \multicolumn{2}{c|}{\textbf{Rot}} &
        \multicolumn{2}{c}{\textbf{Trans}} \\
        & \textbf{$\pm 5^\circ$} & \textbf{$\pm 10^\circ$} 
        & \textbf{$\pm 0.25$m} & \textbf{$\pm 0.5$m} \\
        \midrule
        Initial  & 38.13 & 38.02 & 37.44 & 36.15 \\
        \textbf{Refined} & \textbf{39.43} & \textbf{39.37} & \textbf{39.42} & \textbf{39.34} \\
        \bottomrule
    \end{tabular}
    
\end{table}

\section{Limitations}
First, our cross-state alignment relies on CLIP features, which may fail when multiple objects share similar appearance. This component is not essential and can be replaced with pixel-level or hybrid features to improve robustness. Second, our system does not handle lighting variations, leading to static shadows when objects move. Incorporating relighting into the reconstruction pipeline would enable more realistic, illumination-consistent scene representations.
\section{Conclusion}
We presented a recurrent framework for discrete-state 3D scene fusion that incrementally integrates observations across multiple scene configurations into a single evolving Gaussian representation. Unlike prior methods that either focus on per-state reconstruction or rely on expensive object-centric optimization, our approach models structured scene evolution while maintaining efficiency and scalability. Through CLIP-guided change detection, semantic–geometric pose refinement, and a streaming voxel-based fusion strategy, the system preserves stable structures and incorporates newly revealed regions without reinitialization. Our recurrent Gaussian representation also supports object-level novel state synthesis, demonstrating strong controllability across diverse scene changes.
{
    \small
    \bibliographystyle{ieeenat_fullname}
    \bibliography{main}
}
\appendix
\newpage
\setcounter{page}{1}
\maketitlesupplementary

\section{State fusion}
\begin{figure}[!h]
    \centering
    \includegraphics[width=1\linewidth]{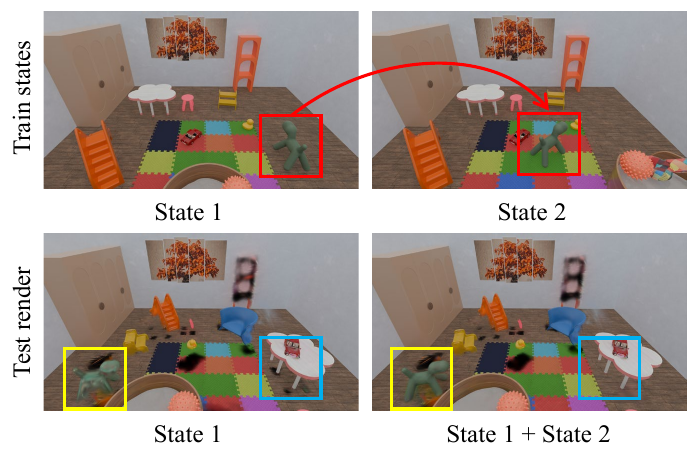}
    \caption{Illustration of state fusion. The top row shows two training states, where the red box highlights the object that changes between states. The bottom row presents renderings of the test state using only State 1 (left) and using the fused representation from State 1 and State 2 (right). The yellow box marks the moved object's foreground in the test state, while the blue box indicates the background region previously occluded by the object.}
    \label{comp}
\end{figure}

Figure~\ref{comp} provides a concrete example demonstrating how our method leverages multiple training states to improve reconstruction quality. As the foreground object moves from State 1 to State 2, our approach fuses the complementary observations captured across states. This leads to two notable improvements in the test-state rendering. First, the background region previously occluded by the object in State 1 becomes visible in State 2, enabling our method to accurately recover the missing background content (blue box). Second, the toy’s right side is insufficiently observed in State 1, resulting in artifacts when rendered in the test state. After incorporating the additional observations from State 2, the reconstructed foreground becomes significantly more complete and artifact-free (yellow box), highlighting the advantages of our fusion strategy.

\begin{figure*}[!ht]
    \centering
    \includegraphics[width=1\linewidth]{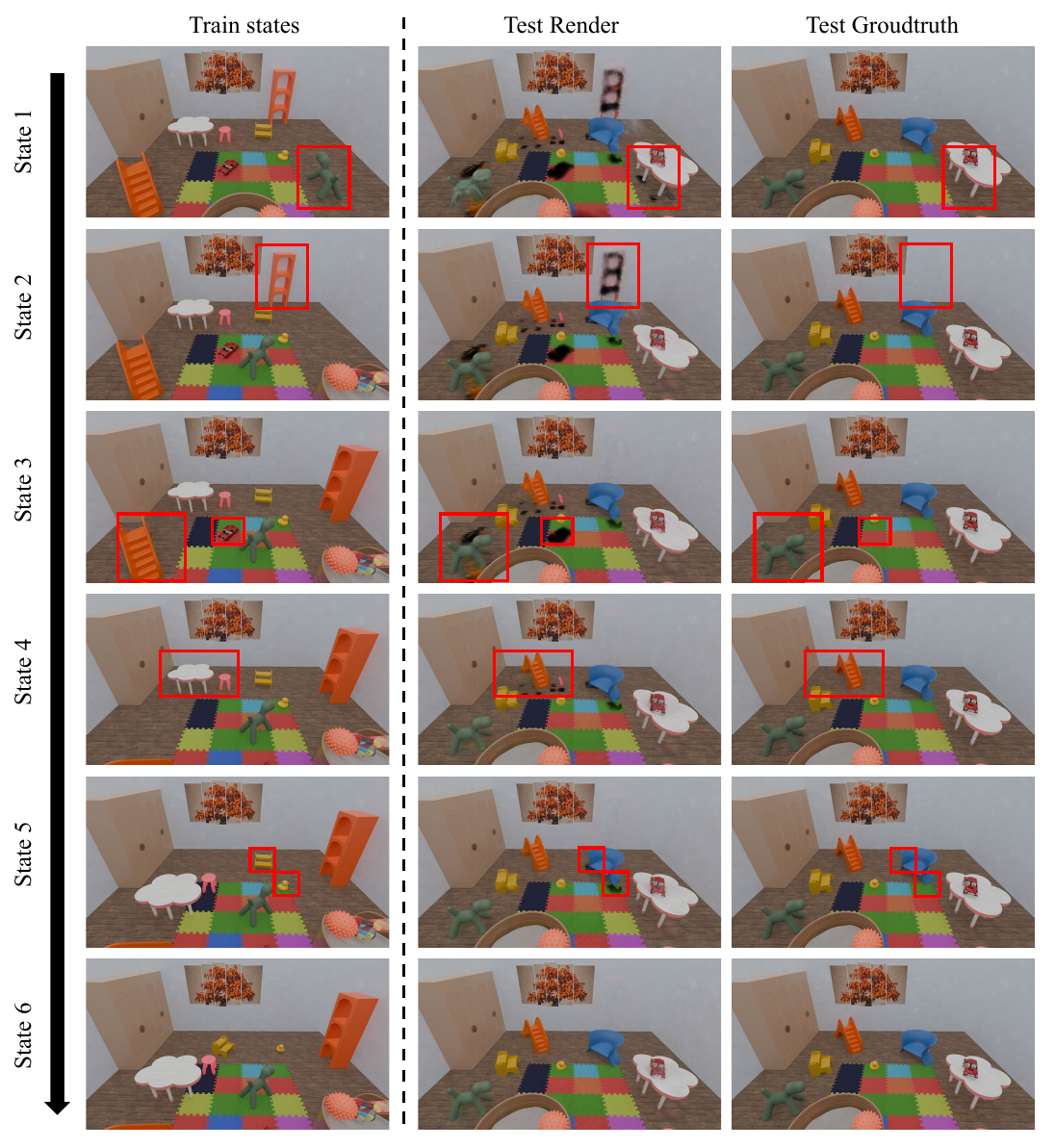}
    \caption{Long-sequence fusion analysis. The left column shows a sequence of training states with progressively changing object configurations, where the red boxes highlight the objects that will move in the subsequent state. The middle column visualizes our method’s test-state rendering as an increasing number of states is incorporated, illustrating how long-sequence information is continuously accumulated and fused to produce increasingly accurate reconstructions. The right column provides the corresponding ground-truth images from the test state.}

    \label{long}
\end{figure*}
\section{Long-sequence fusion}
\begin{figure*}[!t]
    \centering
    \includegraphics[width=1\linewidth]{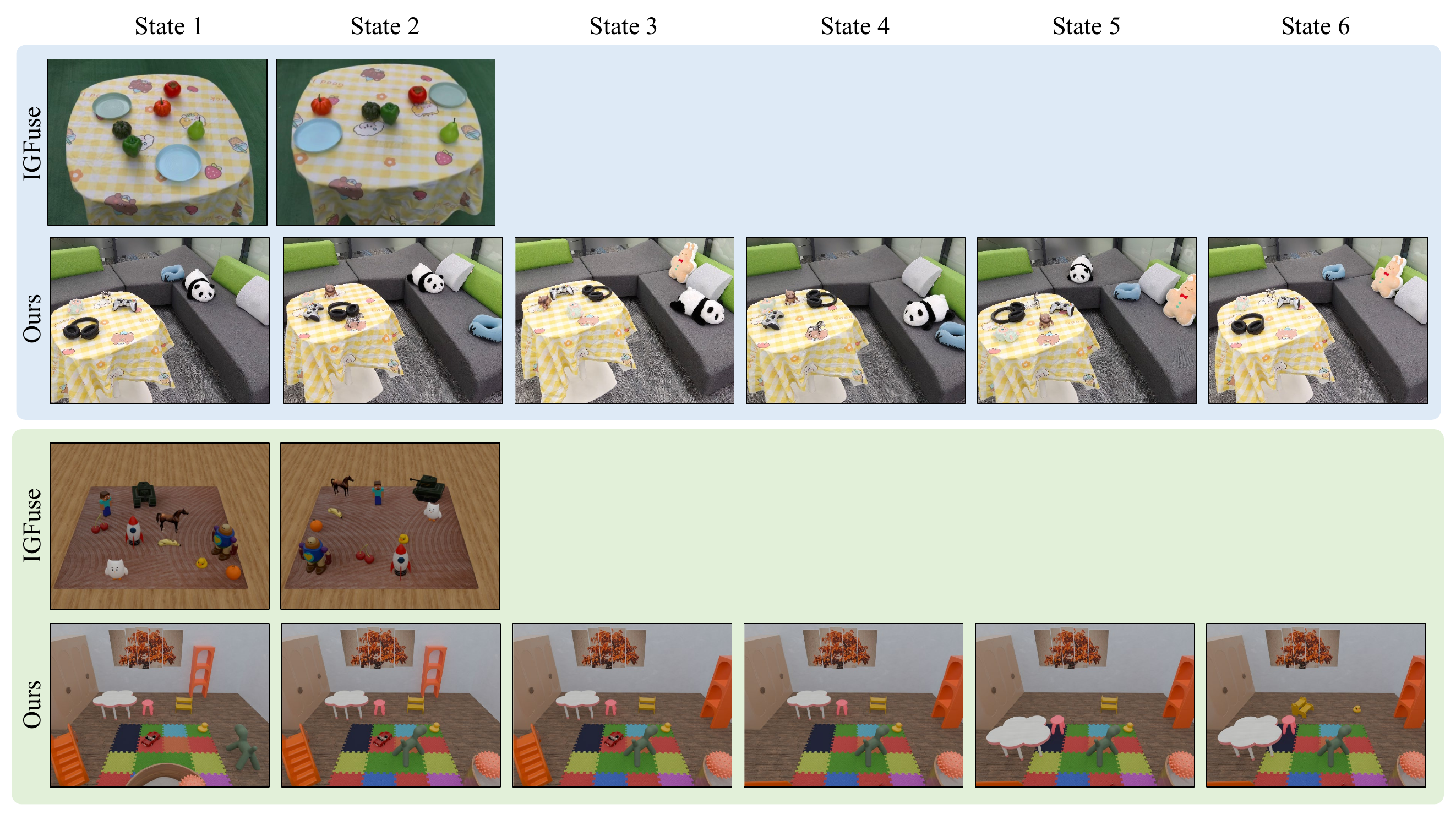}
    \caption{Comparison of state transitions in IGFuse and our long-sequence dataset.}

    \label{data}
\end{figure*}
Building on the short-term state fusion demonstrated previously, we further extend our approach to long-term sequences, as illustrated in Figure~\ref{long}. As more training states are incorporated, our model continuously accumulates complementary observations and progressively disentangles the objects that move across the sequence. This long-term fusion enables both foreground and background regions to be incrementally refined, leading to increasingly accurate test-state renderings. As shown, the reconstructed results gradually converge toward the ground-truth images as the number of fused states increases.

\section{Long-sequence dataset}
Figure~\ref{data} contrasts the interaction patterns underlying IGFuse with those in our long-sequence dataset. IGFuse operates on a short sequence of states in which all objects are required to change simultaneously across state transitions. This design restricts the diversity of scene evolution and limits applicability to highly controlled, synthetic scenarios. In contrast, our dataset provides substantially longer state sequences in which only one or two objects change at each step, and supports natural object addition and removal. Such localized and incremental state evolution is far more representative of real-world scene dynamics, enabling more flexible interaction modeling and allowing our method to accumulate complementary observations over time for consistent long-term reconstruction.

\section{Visibility-Guided Object Refinement}
\begin{figure}[!t]
    \centering
    \includegraphics[width=1\linewidth]{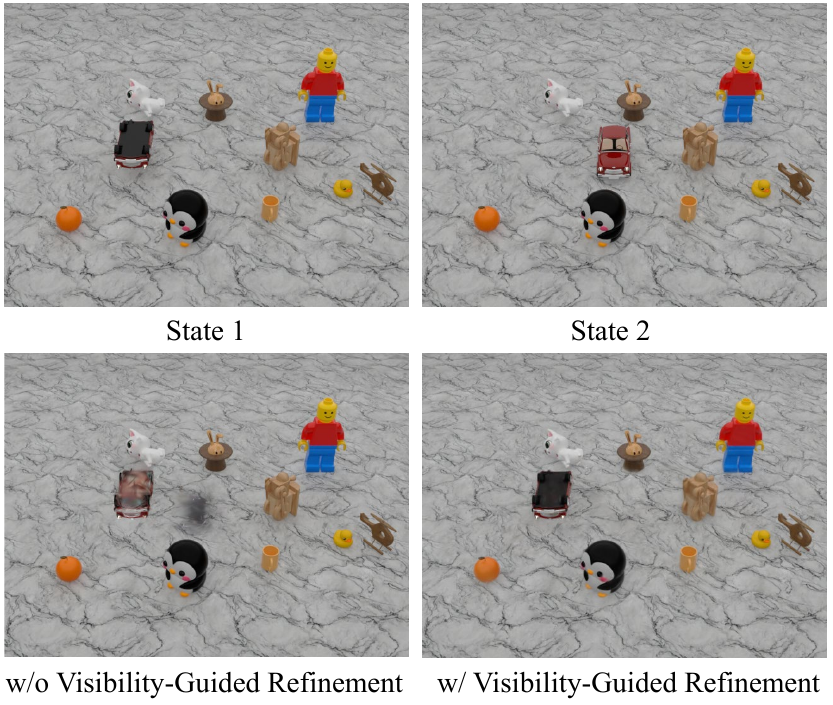}
    \caption{
    Comparison of reconstruction quality when revisiting a previously observed state. 
    We first optimize the Gaussian scene using \textbf{State~1}, then fuse the new observations from \textbf{State~2}, 
    and finally render the scene again from the viewpoints of \textbf{State~1}. 
    The bottom row shows the results \textbf{without} and \textbf{with} our Visibility-Guided Refinement. 
    Note that no recurrent optimization is applied in this experiment.
}
    \label{vis_e}
\end{figure}
We first optimize the scene using State~1, then incorporate new observations from State~2, and finally render the scene again from the viewpoints of State~1 to examine how well the system preserves previously learned content. With Visibility-Guided Refinement, optimization is applied only to Gaussians that are visible in the current state. Consequently, regions that are occluded in State~2, such as the underside of the toy car and the floor beneath it, are not updated. This selective refinement preserves the memory of State~1 and avoids overwriting content that is no longer visible.

In contrast, without Visibility-Guided Refinement, all Gaussians are optimized regardless of visibility. This leads to forgetting, since occluded areas are incorrectly updated using incomplete information from State~2. As illustrated, the underside of the car and the corresponding floor region lose the appearance learned in State~1, showing how naïve full-scene optimization fails to maintain consistency across states.

\section{Hyperparameter}
From Table~\ref{tab:lambda_results}, we observe that the hyperparameter $\lambda_r$ achieves the highest PSNR when set to $0.5$, indicating that this value provides the best balance between preserving information from previous states and incorporating newly observed data. Smaller values under-emphasize historical consistency, while larger values overweight prior states and reduce adaptation to the current observations.

Table~\ref{tab:voxel_results} shows that reducing the voxel size gradually improves PSNR, since finer voxel discretization provides more accurate localization for visibility masking and Gaussian selection. However, when the voxel size decreases to $0.01$, the cost of computing which voxels need to be refined increases dramatically, leading to a significant rise in processing time. Considering the trade-off between reconstruction quality and computational overhead, we adopt a voxel size of $0.05$ in our experiments.

\begin{table}[h]
\centering
\begin{tabular}{c|ccccc}
\toprule
$\lambda_r$ & 0.2 & 0.4 & 0.5 & 0.6 & 0.8 \\
\midrule
PSNR & 
 31.74 & 31.81 & 31.82& 31.62 & 31.40 \\
SSIM &
0.968 & 0.968 & 0.968 & 0.968 & 0.967 \\
\bottomrule
\end{tabular}
\caption{Results under different $\lambda$ values.}
\label{tab:lambda_results}
\end{table}

\begin{table}[h]
\centering
\begin{tabular}{c|cccc}
\toprule
\textbf{Voxel size} & 0.5 & 0.1 & 0.05 & 0.01 \\
\midrule
PSNR & 
31.18 & 31.51 & 31.74 & 32.16 \\
SSIM &
0.964 & 0.967 & 0.968 & 0.969 \\
Time (min) &
3.00 & 3.31 & 3.52 & 42.67 \\
\bottomrule
\end{tabular}
\caption{Results under different voxel sizes.}
\label{tab:voxel_results}
\end{table}

\section{Detail time}

\begin{table}[h]
\centering
\begin{tabular}{l|c}
\toprule
\textbf{Module} & \textbf{Time (min)} \\
\midrule
    Initial Gaussian model & 13.2 \\
    Visual Change Localization & 1.5 \\
    Cross-State Object Association & 0.5 \\
    Geometric Pose Alignment & 7.1 \\
    Visibility-Guided Object Refinement & 3.5 \\
    Recurrent State Optimization & 3.1 \\
\midrule
\textbf{Total} & \textbf{28.9} \\
\bottomrule
\end{tabular}
\caption{Time consumption of different modules.}
\label{tab:module_time}
\end{table}

Table~\ref{tab:module_time} summarizes the runtime of each functional module in our pipeline, using the training process on a short real-world sequence as an example to analyze the sources of computational cost. As shown, training the initial model $G_{\text{rec}}^0$ requires about 13 minutes, representing the dominant portion of the total runtime and serving as the main computational bottleneck. The Visual Change Localization and Cross-State Object Association stages are highly efficient, taking only about 2 minutes in total to complete 2D pixel-level detection and matching. The Geometric Pose Alignment stage takes approximately 7 minutes to estimate the cross-state transformation $T_{\text{fine}}$. In the Visibility-Guided Object Refinement stage, around 3 minutes are devoted to ray tracing to compute $\text{Mask}_{\text{vis}}$. Finally, the Recurrent State Optimization step adds another 3 minutes.


\end{document}